\UseRawInputEncoding
\pdfoutput=1

\documentclass[10pt,twocolumn,letterpaper]{article}

\usepackage[algorithms]{wacv} 

\usepackage{amsmath}
\usepackage{amssymb}
\usepackage{booktabs}

\usepackage{algorithm}
\usepackage{algorithmic}

\usepackage{graphicx}   
\usepackage{subcaption} 

\usepackage{cite}
\usepackage{tcolorbox}
\newtheorem{proposition}{Proposition}
\usepackage{multirow}
\usepackage{mdframed}

\usepackage{microtype}
\usepackage{subcaption}
\usepackage{tabularx}
\usepackage{booktabs} 
\usepackage{subcaption}
\usepackage{setspace}
\usepackage{mathtools}

\usepackage[pagebackref,breaklinks,colorlinks]{hyperref}

\usepackage[capitalize]{cleveref}
\crefname{section}{Sec.}{Secs.}
\Crefname{section}{Section}{Sections}
\Crefname{table}{Table}{Tables}
\crefname{table}{Tab.}{Tabs.}

\def\wacvPaperID{1941} 
\def\confName{WACV}
\def\confYear{2025}

\begin{document}
\title{Multi-View Factorizing and Disentangling: A Novel Framework for Incomplete Multi-View Multi-Label Classification}

\author{
Wulin Xie\\
Guizhou University\\
China\\
\texttt{\small xwl1085930920@gmail.com}
\and
Lian Zhao\\
Guizhou University\\
China\\
\texttt{\small 2816920140@qq.com}
\and
Jiang Long\\
Guizhou University\\
China\\
\texttt{\small longj\_2000@163.com}
\and
Xiaohuan Lu\\
Guizhou University\\
China\\
\texttt{\small xhlu3@gzu.edu.cn}
\and
Bingyan Nie\thanks{Corresponding author: bingyannie966@gmail.com}\\
Guizhou University\\
China\\
\texttt{\small bingyannie966@gmail.com}
}
\maketitle

\begin{abstract}
Multi-view multi-label classification (MvMLC) has recently garnered significant research attention due to its wide range of real-world applications. However, incompleteness in views and labels is a common challenge, often resulting from data collection oversights and uncertainties in manual annotation. Furthermore, the task of learning robust multi-view representations that are both view-consistent and view-specific from diverse views still a challenge problem in MvMLC. To address these issues, we propose a novel framework for incomplete multi-view multi-label classification (iMvMLC). Our method factorizes multi-view representations into two independent sets of factors: view-consistent and view-specific, and we correspondingly design a graph disentangling loss to fully reduce redundancy between these representations. Additionally, our framework innovatively decomposes consistent representation learning into three key sub-objectives: (i) how to extract view-shared information across different views, (ii) how to eliminate intra-view redundancy in consistent representations, and (iii) how to preserve task-relevant information. To this end, we design a robust task-relevant consistency learning module that collaboratively learns high-quality consistent representations, leveraging a masked cross-view prediction (MCP) strategy and information theory. Notably, all modules in our framework are developed to function effectively under conditions of incomplete views and labels, making our method adaptable to various multi-view and multi-label datasets. Extensive experiments on five datasets demonstrate that our method outperforms other leading approaches.
\end{abstract}

\section{Introduction}
\label{sec:intro}
Nowadays, multi-view data, collected from diverse sources such as SIFT, Gist, and HSV or obtained from various feature extractors, has garnered significant research attention due to its potential to provide richer and more diverse descriptions of samples \cite{Ling2023DualLG}\cite{Xu2024ReliableCM}\cite{Wan2023AutoweightedMC}\cite{diao2023ft2tf}\cite{CMSCGC}. This leads to an intriguing research question: how can multi-view features be effectively integrated to extract meaningful information, many multi-view learning methods have been proposed to address this question\cite{Tan2023SamplelevelMG}\cite{SSGCC}. Concurrently, multi-label classification has long been a crucial area in pattern recognitione\cite{10.1609/aaai.v37i8.26190}\cite{Kobayashi2023forTM}\cite{diao2024learning} as single-label data often fail to capture the complexities of real-world scenarios. For instance, an image might be described by multiple elements such as “neon signs", “sidewalks", and “passing cars", each contributing to a comprehensive understanding of the scene. Consequently, the field of multi-view learning now faces a new composite challenge: multi-view multi-label classification (MvMLC). 

However, most MvMLC methods assume that all views and labels are complete, which does not consistently align with real-world scenarios. For example, if an online article contains only text and images, the option to view a video will not be available. Additionally, manual annotation may overlook certain tags due to errors or budget constraints, reducing the effectiveness of multi-label supervision. This challenge highlights the necessity of incomplete multi-view multi-label classification (iMvMLC), which focuses on scenarios where some views or labels are missing. To address this problem, several related works have been proposed in recent years. For example, Li et al. proposed NAIM3L that leveraged consensus and structural information among labels to ease the incompleteness problem.\cite{9447974} Liu et al. proposed DICNet, a deep instance-level contrastive network that sings stacked autoencoders for view-specific feature extraction and an instance-level contrastive learning scheme to enhance consensus representation.\cite{Liu2023DICNetDI} Liu et al. introduced LMVCAT which leverages a transformer-based network to capture the relationship between different views and labels.\cite{Liu2023IncompleteMM} However, most existing iMvMLC methods rely on end-to-end frameworks to simultaneously learn consistent and view-specific representations. This approach not only creates a Min-Max game that risks converging on sub-optimal solutions in the absence of supplementary information but also significantly increases computational demands as the number of views grows\cite{Ke2024RethinkingMR}. Additionally, coupling view-specific and view-shared information within the same feature space inevitably leads to redundancy between the two. To ensure simultaneous learning of these different types of information during representation learning, models inevitably discarded some specific semantic details during the encoding process, resulting in the loss of some view-specific information.

To address these challenges, we propose a multi-view factorizing and disentangling framework (MVFD) for the iMvMLC task. Unlike existing approaches that utilize an end-to-end framework to concurrently learn consistent and view-specific representations, our method factorizes multi-view representations into view-consistent and view-specific factors and employs a two-stage multi-view learning paradigm to learn each separately. Furthermore, to obtain pure view-consistent representation for downstream task,  we decompose consistent representation learning into three objectives: extracting view-shared information, eliminating redundant intra-view information, and preserving task-relevant information. We correspondingly design a collaborative consistency learning module based on a masked cross-view prediction (MCP) strategy and information theory. Finally, we design a graph disentangling loss to fully reduce redundancy between view-consistent and view-specific representations.

In summary, the contributions of our method can be highlighted as follows:

\begin{itemize}
\item We propose a novel framework for iMvMLC that factorizes multi-view representations into view-consistent and view-specific factors and employs a two-stage learning paradigm to learn them separately. Additionally, our framework is compatible with arbitrary incomplete multi-view and weak multi-label data.

\item Our framework innovatively decomposes the consistency learning task into three sub-objects and combines the masked cross-view prediction (MCP) strategy and information theory for collaborative optimization.

\item We design a graph disentangling loss aimed at minimizing redundant information between view-consistent and view-specific representations, which can also be adapted to current multi-view representation learning methods. Extensive experiments on five datasets validate that our method outperforms other state-of-the-art approaches for the iMvMLC task.
\end{itemize}

\section{Related Works}
\subsection{Multi-View Multi-Label Classification}
In recent years, many works for multi-view multi-label classification (MvMLC) tasks have been proposed\cite{10.5555/2886521.2886708}\cite{10.1016/j.neucom.2019.09.009}\cite{9939043}. Sun et al. introduce the Latent Conditional Bernoulli Mixture (LCBM), a multi-view probabilistic model for multi-label classification that enhances label dependency modeling and generalization through a novel integration of Bernoulli mixture models and a Gaussian Mixture Variational Autoencoder (GMVAE) within a variational inference framework\cite{9000714}. Zhang et al. proposed LSA-MML, which utilizes matrix factorization and the Hilbert-Schmidt Independence Criterion to align and enhance the representation of multiple views. \cite{10.5555/3504035.3504576}. In addition, neural networks (DNN) have also been introduced in the MvML task. For example, Wu et al. proposed SIMM that combines confusion adversarial learning and the orthogonal constraint to effectively integrate and optimize both shared and view-specific information \cite{10.5555/3367471.3367581}. Zhao et al. proposed CDMM that addresses the challenge of maintaining consistency and diversity among multiple views by using separate classifiers for each view, integrating the Hilbert–Schmidt Independence Criterion for diversity, and incorporating label correlation and view contribution factors to enhance prediction accuracy.\cite{Zhao2021ConsistencyAD}

\subsection{Incomplete Multi-View Multi-Label Classification}
However, existing MvMLC methods ignore the incompleteness problem in MvMLC and can't handle missing views or missing labels. To this end, many methods for IMvMLC have been developed that can handle missing views and missing labels simultaneously\cite{nie2024incompletemultiviewmultilabelclassification,Xie2024UncertaintyAwarePA,Long2024MultiscaleLP,Lu2024TaskAugmentedCI}. For instance, Li et al. proposed NAIM3L that effectively handles missing labels, and incomplete and non-aligned views with minimal assumptions, employing a single hyper-parameter and a high-rank modeling approach for global label structures\cite{9447974}. Tan et al. proposed iMVWL which simultaneously addresses incomplete views and weak labels by learning a shared subspace that captures cross-view relationships and local label correlations\cite{Tan2018IncompleteMW}.  Recently, DNN has proven its huge potential in iMvMLC tasks. Liu et al. proposed DICNet that introduces a DNN-based framework for iMvMLC that effectively handles incomplete data through view-specific autoencoders\cite{Liu2023DICNetDI}. Liu et al. proposed MTD that enhances consistency and uniqueness in view representations through a two-channel encoder system and utilizes a weak label-guided graph regularization to improve feature decoupling and preserve geometric structures in embeddings\cite{liu2024masked}.

\section{Method}
In this section, we introduce our method in detail from the following three aspects, namely factorized consistent representation Learning, view-specific representation disentangling, and multi-label classification. For the convenience of description, we first introduce the formal problem definition and common notations.

\subsection{Problem definition}
 In order to clearly describe the iMvMLC task, we define original multi-view data as $\left \{ \mathbf{X}^{(v)} \in \mathbb{R} ^{n\times d_{v} }  \right \}_{v=1}^{m} $
  where $m$, $n$, $d_{v}$ denotes the number of views, sample, and the original dimension of $v$-th view feature, respectively. $\mathbf{Y}\in \left \{ 0,1 \right \}^{n\times c}$ denotes the label matrix with $c$ categories and $\mathbf{Y}_{i,j}=1$ denotes that the $i$-th sample belongs to the $j$-th category. Furthermore, we introduce missing view indicator $\mathbf{W}\in \left \{ 0,1 \right \} ^{n\times m} $ and missing label indicator $\mathbf{G}\in \left \{ 0,1 \right \} ^{n\times m}$, respectively, where $\mathbf{W}_{i,j} =1$ means the $j$-th view of $i$-th sample is available, otherwise $\mathbf{W}_{i,j} =0$. Similarly, $\mathbf{G}_{i,j} =1$ represents $i$-th sample's $j$-th label is known, otherwise $\mathbf{G}_{i,j} =0$. Additionally, we filled the missing data in $\left \{ \mathbf{X}^{(v)} \in \mathbb{R} ^{n\times d_{v} }  \right \}_{v=1}^{m} $ and $\mathbf{Y}\in \left \{ 0,1 \right \}^{n\times c}$ with '0' in the data pre-processing step. The target of our method is to train a network that can accurately predict the categories of incomplete multi-view data.

\begin{figure*}[ht]
    \centering
    \includegraphics[width=1.0\textwidth]{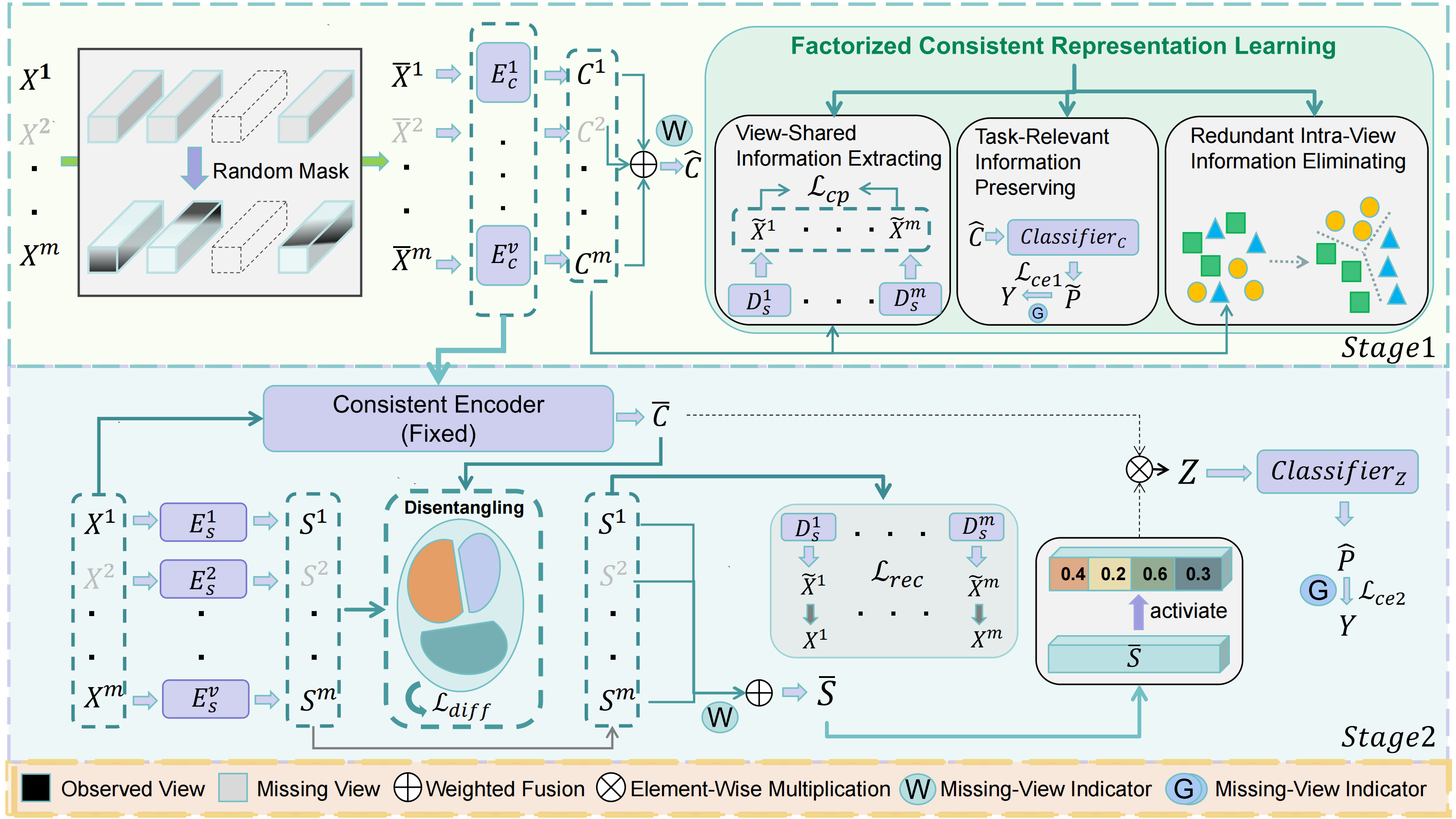}
    \caption{An overview of our MVFD framework. In the first age, we first randomly masked fragments of input features. Then we factorize consistent representation learning into three sub-goals for collaborative optimization. In the second stage, we freeze consistent encoders trained in the first stage and leverage learned consistent representation and our graph disentangling loss to guide the disentanglement process of view-specific information.  }
    \label{fig:enter-label}
\end{figure*}

\begin{figure}[!t]
    \centering
    \includegraphics[width=1.0\linewidth]{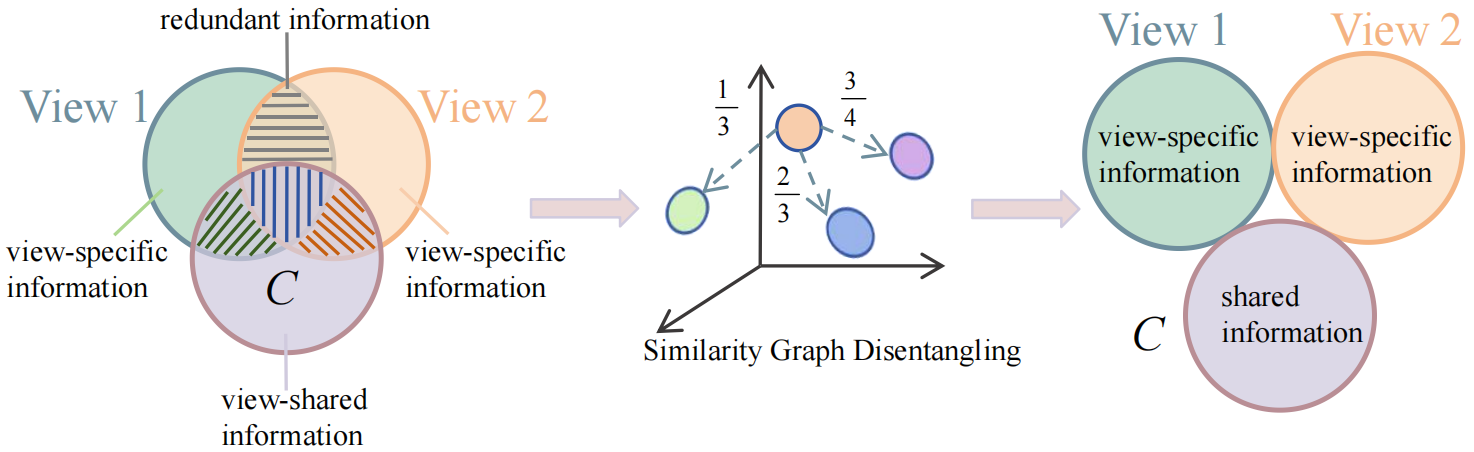}
    \caption{We define a complete multi-view data contain view-specific information, view-shared information, and redundant information between. The goal of our graph disentangling loss is fully eliminate redundancy between them and obtain disentangled view-specific information.}
    \label{fig:your-label}
\end{figure}

\subsection{Factorized Consistent Representation Learning}
\textit{\textbf{View-shared information extracting}}: As we know, the desired latent consistent representation should encode the shared information from observed views\cite{sunprogram}\cite{sun2024ultra}\cite{cheng2024spt}. As discussed in \cite{9258396} and \cite{Ke2024RethinkingMR}, an effective view-consistent representation can concurrently reconstruct diverse views of the same object if each view is conditionally independent given the shared multi-view representation. Thus, we define the desired multi-view consistent representation as follows:
\begin{tcolorbox}[colframe=black, colback=white, boxrule=0.5mm, arc=2mm]
\begin{proposition}
A multi-view consistent representation $c$ contains the shared information across different views if each observation, i.e., $x^{(v)} $ from ${x^{(1} ,..., x^{(m)} }$,can be reconstructed from a mapping $f^{(v)}\left ( \cdot  \right ) $, i.e., $x^{(v)} = f^{(v)}\left ( c \right ) .$
\end{proposition}
\end{tcolorbox}

where $c$ denotes the consistent representation of multiple views. Additionally, since the original multi-view data contains redundant view-specific information that may interfere with consistency learning, we are inspired by masked autoencoders (MAE) \cite{He2021MaskedAA}, which randomly mask image patches. As shown in Fig. 1, in the first stage, we apply a random masking technique to the input multi-view vector data. Specifically, we first construct a matrix $\mathbf{M}^{(v)} \in \left\{ 0,1 \right\}^{n\times d_{v}}$ for any view $v$, where all elements are initially set to $1$. For each row $m_{i,\colon}^{(v)} \in \mathbf{M}^{(v)}$, where $i \in \left\{ 1,2,\dots,n \right\}$, we randomly select an initial point $p_{i}^{(v)} \in \left[ 0,d_{v} - l^{(v)} \right]$ and set the elements to $0$ from position $p_{i}^{(v)}$ to $p_{i}^{(v)} + l^{(v)}$, where $l^{(v)}$ denotes the mask length in view $v$. Specifically, $l^{(v)} = d_{v} \cdot \delta$, where $\delta$ is a parameter representing the mask ratio. Finally, we obtain the weight matrix $\mathbf{\bar{M}}^{(v)}$, and use the weight matrices $\left\{\mathbf{\bar{M}}^{(v)}\right\}_{v=1}^{m}$ to mask the original multi-view data as follows:

\begin{equation}
\left \{ \mathbf{\bar{X}}^{(v)}  \right \} _{v = 1}^{m} = \left \{ \mathbf{X}^{(v)} \odot\mathbf{ \bar{M}}^{v}  \right \} _{v = 1}^{m}
\end{equation}
where $ \odot$ denotes Hadamard product and $\left \{ \mathbf{\bar{X}}^{(v)}  \right \} _{v = 1}^{m}$ is final masked multi-view data. Then we use a set of multi-layer perceptrons (MLP) as consistent encoders to extract shared information from multiple views: $\left \{E_{c}^{(v)}: \mathbf{\bar{X}}^{(v)} \in \mathbb{R}^{n\times d_{v} }   \to \mathbf{C}^{(v)} \in \mathbb{R} ^{n\times d_{e} }       \right \} _{v=1}^{m} $ where $E_{c}^{(v)}$ means the consistent encoder of view $v$,  $\mathbf{C} ^{(v)}\in \mathbb{R} ^{n\times d_{e} } $ denotes the $v$-th view's embedding consistent feature and $d_{e}$ represents the embedding dimension. Then we obtain comprehensive consistent representation $\left \{ \mathbf{\hat{C}} _{i,:} \right \}_{i=1}^{n}  $ as follows:
\begin{equation}
    \mathbf{\hat{C}} _{i,:} =\sum_{v=1}^{m} \frac{\mathbf{C} _{i,:}^{(v)} \mathbf{W}_{i,v} }{ {\textstyle \sum_{v}^{}} \mathbf{W}_{i,v}}
\end{equation}
where $\mathbf{W}\in \mathbb{R} ^{n\times m}  $ means the missing view indicator matrix. To satisfy the coniditon in proposition 1, we utilize a set of MLPs to construct view-specific decoders as mapping functions and put consistent embedding features as input to reconstruct raw multi-view data as follows: $\left \{D_{c}^{v}: \mathbf{\hat{C} } \in \mathbb{R}^{n\times d_{e} }   \to \mathbf{X^{'} }^{(v)} \in \mathbb{R} ^{n\times d_{v} } \right \} _{v=1}^{m} $ where $\mathbf{X^{'} }^{(v)}$ denotes the reconstructed feature of view $v$. Then we design masked consistent prediction loss as follows:
\begin{equation}
 \ell_{cp} =  \frac{1}{m} \sum_{v=1}^{m}  \ell_{cp}^{(v)} = \frac{1}{m} \sum_{v=1}^{m} \left ( \frac{1}{d_{v}}\sum_{i=1}^{n} \left \| \mathbf{X'} ^{(v)}_{i}  -\mathbf{X} ^{(v)}_{i}   \right \|^{2}_{2}  W_{i,v}   \right ) 
\end{equation}
where $W_{i,v}$ denotes the availability of the $v$-th view in $i$-th sample. By minimizing Eq.(3), $\mathbf{\hat{C} }$ can encodes shared information from available views, and different samples (regardless of their missing
patterns) are associated with representations in a common space.

\textit{\textbf{Redundant intra-view information eliminating}}: While minimizing Eq.(3) can extract consistency among different views from high-level features, the reconstruction objective inevitably results in the same features
to maintain the view-private information for individual view\cite{Xu2021MultilevelFL}, which hinders the extraction of consistent information. To this end, we introduce contrastive learning into semantic space to eliminate view-specific information within high-level consistent features. Specifically, we first obtain predictions of all views through a shared classifier as follows: $   \left \{\Psi _{c}: \mathbf{C}^{(v)}  \in \mathbb{R}^{n\times d_{e} }   \to \mathbf{\bar{P} }^{(v)} \in \mathbb{R} ^{n\times c } \right \} _{v=1}^{m}$ where $\Psi _{c}$ denotes the shared classifier and $\mathbf{\bar{P} }^{(v)}_{i,j} $ means the confidence of sample $i$ belongs to category $j$ in the $v$-th view. In real-world situations, different views of the same instance may have different predictive confidence for the same category due to interference with view-specific information. To eliminate view-specific information and obtain pure consistent representation, we need to maintain semantic consistency between different views of the same instance. In other words, predictions from different views of the same sample should be consistent. To this end, we regard predictions as semantic features and propose semantic contrastive loss as follows:
\begin{equation}
\begin{aligned}
    \ell _{sc} &=\frac{1}{2}\sum_{v=1}^{m} \sum_{u\ne v}^{}\ell_{sc}^{(vu)} =  \frac{1}{2}\sum_{v=1}^{m} \sum_{u\ne v}^{} \\
    &\left ( \sum_{i=1}^{n} log\frac{exp\left ( cos\left ( \mathbf{\bar{P}}_{i,:}^{(v)}, \mathbf{\bar{P}}_{i,:}^{(u)}  \right ) / \tau  \right )  }{ {\textstyle } {\textstyle exp\left ( cos\left ( \mathbf{\bar{P}}_{i,:}^{(v)}, \mathbf{\bar{P}}_{i,:}^{(u)}  \right ) / \tau  \right ) + \mathbb{S} _{neg}   }  }  \right )
\end{aligned}
\end{equation}
where $\mathbb{S} _{neg}  = {\textstyle \sum_{r=u,v}^{}}  {\textstyle \sum_{j=1,j\ne i}^{n}} exp\left ( cos\left ( \mathbf{\bar{P}}_{i,:}^{(v)}, \mathbf{\bar{P}}_{j,:}^{(u)}  \right ) / \tau  \right ) $ and function $cos\left ( a,b \right ) $ means the cosine similarity between embedding feature $a$ and $b$. In Eq.(4), for any anchor instance $\mathbf{\bar{P}}_{i,:}^{(v)}$, we have $(N-1)$ positive instances: $\bar{\mathbf{P}} _{i,:}^{(u)}| _{u\ne v} $ and $N(M-1)$ negative instances: $\bar{\mathbf{P}} _{j,:}^{(v)}| _{j\ne i} $. Through minimizing Eq.(4), we can learn semantic consistency and eliminate view-specific information within consistent representation.

\textit{\textbf{Task-relevant information preserving}}: Notably, the minimal sufficient representation obtained in contrastive learning is insufficient for downstream tasks since it eliminates task-relevant non-shared information\cite{Wang2022RethinkingMS}. To cope with this issue, inspired by \cite{Tishby2000TheIB},  we can obtain task-relevant consistent representation by minimizing as follows:
\begin{equation}
    \min_{\hat{C} }  -I\left ( \hat{\mathbf{C}} ;\mathbf{Y} \right ) 
\end{equation}
where $I\left ( \hat{\mathbf{C}} ;\mathbf{Y} \right ) $ means the mutual information between consistent representation $\hat{\mathbf{C}}$ and label $\mathbf{Y}$ and minimizing Eq.(5) can be regarded as minimizing the common cross-entropy loss\cite{pmlr-v130-lee21a}. To achieve this goal, we take $\Psi _{c}$ as classifier and put $\hat{C} $ in it to obtain predictions $\tilde{\mathbf{P} }  = \Psi _{c} \left ( \hat{\mathbf{C}}  \right ) $ where $\tilde{\mathbf{P} } \in \mathbb{R}^{n\times c}  $. Then we minimize cross-entropy loss as follows:
\begin{equation}
\begin{aligned}
\ell _{ce1} =& \frac{1}{ {\textstyle \sum_{i,j}^{}\mathbf{G}_{i,j} } } \sum_{i=1}^{n} \sum_{j=1}^{c}\bigg( (\mathbf{Y}_{i,j}\log_{}{\left ( \tilde{\mathbf{P} } _{i.j}   \right ) } \\
&+\left ( 1-\mathbf{Y}_{i,j}  \right )\log_{}{\left ( 1- \tilde{\mathbf{P} } _{i.j}   \right ) }  \bigg) \mathbf{G}_{i,j}  
\end{aligned}
\end{equation}
where $\mathbf{G}$ denotes the missing-label indicator. Overall, all loss functions in the second stage becomes:
\begin{equation}
    \ell _{stage1} = \ell _{ce1} + \alpha \ell _{cp} + \beta \ell _{sc} 
\end{equation}

\subsection{View-Specific Representation Disentangling}
View-specific information encodes unique semantic details within a view—such as angles, distances, and positional data. A complete view combines both view-specific and view-consistent information through complex higher-order interactions\cite{Ke2024RethinkingMR}. This section discusses how to disentangle these two types to obtain pure view-specific information.

An intuitive approach is to use the view consistency factor we have learned to disentangle the view-specific factor, thereby removing redundant information from the latter. Specifically, we first introduce an autoencoder to extract coarse view-specific features from the raw multi-view data: $\left \{ E_{s}^{(v)}: \mathbf{X}^{(v)} \in \mathbb{R}^{n\times d_{v} } \to \mathbf{S}^{(v)} \in \mathbb{R} ^{n\times d_{e} } \right \} _{v=1}^{m}$, where $E_{s}^{(v)}$ from $E_{s}^{(1)},...,E_{s}^{(m)}$ is a set of view-specific encoders constructed by MLPs, and $\mathbf{S}^{(v)}$ represents the high-level view-specific feature of the $v$-th view. We then utilize a set of view-specific decoders to reconstruct the raw multi-view data: $\left \{ D_{s}^{(v)}: \mathbf{S}^{(v)} \in \mathbb{R}^{n\times d_{v} } \to \mathbf{\hat{X} }^{(v)} \in \mathbb{R} ^{n\times d_{e} } \right \}_{v=1}^{m}  $. The reconstruction loss is calculated as follows:

\begin{equation}
\ell_{rec} =\frac{1}{m} \sum_{v=1}^{m}\ell_{rec}^{(v)} =\frac{1}{m} \sum_{v=1}^{m}\left ( \frac{1}{d_{v} }\sum_{i=1}^{n}\left \|  \mathbf{\hat{X}}_{i}^{(v)}- \mathbf{X}_{i}^{(v)} \right \|_{2}^{2}   W_{i,v} \right )
\end{equation}

where $W$ is the missing-view indicator, introduced to mitigate the negative effects of missing views in calculation.

Following this, the next goal is how to leverage learned consistent representation to distill pure view-specific information from coarse view-specifc feature. To this end, we firstly freeze the consistent encoders trained in the first stage. We extract shared multi-view features: $\left \{E_{c(fixed)}^{(v)}: \mathbf{X}^{(v)} \in \mathbb{R}^{n\times d_{v} }   \to \mathbf{C'}^{(v)} \in \mathbb{R} ^{n\times d_{e} }       \right \} _{v=1}^{m}$. Then we can obtain a comprehensive consistent representation as follow:
\begin{equation}
    \mathbf{\bar{C}} _{i,:} =\sum_{v=1}^{m} \frac{\mathbf{C'} _{i,:}^{(v)} \mathbf{W}_{i,v} }{ {\textstyle \sum_{v}^{}} \mathbf{W}_{i,v}}
\end{equation}
Then we propose our graph disentangling loss to eliminate redundancy between different types of information as follow:
\begin{equation}
\begin{aligned}
    \ell_{gd}  = &\sum_{i=1}^{n} \Bigg ( \frac{1}{ {\textstyle \sum_{v}^{}W_{i,v} } } \sum_{v=1}^{m} \mathcal{S}\left (\mathbf{ \bar{C}}_{i}, \mathbf{S}_{i}^{(v)} \right )W_{i,v} \\
   &+\frac{1}{ {\textstyle \sum_{u,v}^{}W_{i,u}W_{i,v}  } } \sum_{v=1}^{m}\sum_{u\ne v}^{m}\mathcal{S}\left (\mathbf{S}_{i}^{(v)} , \mathbf{S}_{i}^{(u)} \right )W_{i,v}W_{i,u}\Bigg ) 
\end{aligned}
\end{equation}
where $\mathcal{S}$ is the cosine similarity function and $\mathcal{S}\left ( a,b \right ) = \frac{a^{T}b }{\left \| a \right \|_{2}\cdot   \left \| b \right \|_{2}  }$. Equation (10) comprises two terms: the first minimizes the similarity between consistent information and view-specific information, and the second minimizes the similarity between view-specific information from different views. Specifically, $W_{i,v} $ in the first term ignores missing views, and $W_{i,v}W_{i,u}$ in the second term identifies instance pairs present in both views. By minimizing these similarities, we indirectly obtain disentangled representations while maintaining clear distinctions between view-specific features and shared features. This allows us to naturally perform weighted cross-view fusion to obtain a unique view-specific representation as follows:
\begin{equation}
    \mathbf{\bar{S}} _{i,:} =\sum_{v=1}^{m} \frac{\mathbf{S} _{i,:}^{(v)} \mathbf{W}_{i,v} }{ {\textstyle \sum_{v}^{}} \mathbf{W}_{i,v}}
\end{equation}

\subsection{Multi-Label Classification}
Up to now, we have obtained the consistent representation $\bar{\mathbf{C}}$ and the disentangled view-specific representation $\bar{\mathbf{S}}$. The next challenge is how to combine them to derive the final discriminative feature for classification. Unlike previous works that rely on summation or concatenation operations, we are inspired by \cite{ma2024rewrite} and \cite{liu2024masked}, which have demonstrated the effectiveness of the star operation (element-wise multiplication) in capturing subtle differences and complex patterns between data. Thus, we leverage the star operation to enhance the shared information via the view-specific information as follows:
\begin{equation}
    \mathbf{Z}  = Sigmoid\left ( \bar{\mathbf{S} }  \right ) \odot \bar{\mathbf{C} } 
\end{equation}
where $Sigmoid\left ( \cdot  \right )$ is a activate function and $\mathbf{Z} \in \mathbb{R} ^{n\times d_{e} } $ is final discriminative feature. Finally, we can obtain the final prediction $\hat{\mathbf{P} }  = \Psi _{Z} \left ( \mathbf{Z}  \right ) $ where $\hat{\mathbf{P} } \in \mathbb{R}^{n\times c} $ and calculate multi-label classification loss as follows:
\begin{equation}
\begin{aligned}
    \ell _{ce2} =& \frac{1}{ {\textstyle \sum_{i,j}^{}\mathbf{G}_{i,j} } } \sum_{i=1}^{n} \sum_{j=1}^{c}\bigg( ( \mathbf{Y}_{i,j}\log_{}{\left ( \hat{\mathbf{P} } _{i.j}   \right ) } \\
&+\left ( 1-\mathbf{Y}_{i,j}  \right )\log_{}{\left ( 1- \hat{\mathbf{P} } _{i.j}   \right ) }  \bigg) \mathbf{G}_{i,j}  
\end{aligned}
\end{equation}
where $G$ denotes missing-label indicator which introduced to ﬁlter invalid missing labels during calculate process.

Overall, all loss functions in the second stage becomes:
\begin{equation}
    \ell _{stage2} = \ell _{ce2} + \gamma \ell _{rec} + \lambda \ell _{gd}  
\end{equation}
and the training process is described in Algorithm 1 in supplementary materials

\section{Experiments}

\subsection{Experimental Settings}
\textit{\textbf{Datasets and evaluation metrics}}:
Consistent with existing iMvMLC methods\cite{Liu2023DICNetDI}\cite{9447974}\cite{Liu2019LateFI}, we adopt five commonly datasets, i.e., Corel5k\cite{Duygulu2002ObjectRA}, Pascal07\cite{Everingham2014ThePV}, ESPGame\cite{Ahn2004LabelingIW}, IAPRTC12\cite{Grubinger2006TheIT}, and MIRFLICKR\cite{Huiskes2008TheMF}. Additionally, six distinct types of features are extracted as six views from these datasets, namely GIST, HSV, HUE, LAB, RGB, and SIFT. Following previous works\cite{Liu2023DICNetDI}, we select six metrics to evaluate all comparison methods,  i.e., Average Precision (AP), Ranking Loss (RL), the adapted area under the curve(AUC), OneError (OE), and Coverage (Cov). Specifically, for all metrics, higher values indicate better performance.

\textit{\textbf{Incomplete multi-view multi-label data processing}}:
To simulate a real-scenario of missing data, similar to existing works\cite{Tan2018IncompleteMW}\cite{Liu2023DICNetDI}, we construct incomplete multi-view multi-label datasets as follows: for each view, we randomly disable $50\% $instances and replace them with value '$0$' and ensuring that each sample has at least one available view. For each category, we randomly eliminated $50\% $of the positive and negative tags and replaced them with '$0$'. Finally, we randomly select $50\%$ of samples with missing views and missing labels as the training set.

\textit{\textbf{Comparison methods}}:
In order to evaluate the performance of our method, we compare ten top methods, i.e., GLOCAL\cite{8233207}, CDMM\cite{Zhao2021ConsistencyAD}, DM2L\cite{Ma2020ExpandGS}, LVSL\cite{9939043}, iMVWL\cite{Tan2018IncompleteMW}, NAIM3L\cite{9447974}, DICNet\cite{Liu2023DICNetDI}, LMVCAT\cite{Liu2023IncompleteMM}, MTD\cite{liu2024masked} and AIMNet\cite{liu2024attention} with our MVFD on five incomplete multi-view multi-label datasets. Specifically, not all methods for the comparison are suitable for the iMvMLC task, such as GLOCAL, CDMM, DM2L and LVSL, where GLOCAL and DM2L  can deal with incomplete labels and CDMM and LVSL are the MvMLC methods unable to handle any data missing problem. To implement the above four methods on the iMvMLC tasks, we used mean imputation for missing views and regarded the unknown label as the negative tag for methods that are unable to handle partial multi-label data.

\begin{table*}[h]
\centering
\caption{Experimental results of eleven methods on the five datasets with 50\% missing-view rate 50\% missing-label rate, and 70\% training samples (the bottom right digit is the standard deviation).}
\resizebox{\textwidth}{!}{
\begin{tabular}{lcccccccccccc}
\toprule
Data & Metric & GLOCAL & CDMM & DM2L & LVSL & iMVWL & NAIM3L & DICNet & LMVCAT & AIMNet & MTD & \textbf{MVFD} \\
\midrule
\multirow{6}{*}{Corel5k} 
& AP & 0.285$_{0.004}$ & 0.354$_{0.004}$ & 0.262$_{0.005}$ & 0.342$_{0.004}$ & 0.283$_{0.008}$ & 0.309$_{0.004}$ & 0.381$_{0.004}$ & 0.382$_{0.004}$ & 0.400$_{0.010}$ & 0.415$_{0.008}$ & \textbf{0.441$_{0.003}$} \\
& 1-HL & 0.987$_{0.000}$ & 0.987$_{0.000}$ & 0.979$_{0.001}$ & 0.978$_{0.000}$ & 0.979$_{0.001}$ & 0.988$_{0.000}$ & 0.988$_{0.000}$ & 0.986$_{0.000}$ & 0.988$_{0.000}$ & 0.980$_{0.000}$ & \textbf{0.988$_{0.000}$} \\
& 1-RL & 0.840$_{0.003}$ & 0.854$_{0.003}$ & 0.843$_{0.002}$ & 0.850$_{0.005}$ & 0.878$_{0.002}$ & 0.882$_{0.004}$ & 0.882$_{0.003}$ & 0.880$_{0.002}$ & 0.902$_{0.004}$ & 0.893$_{0.004}$ & \textbf{0.903$_{0.003}$} \\
& AUC & 0.843$_{0.003}$ & 0.886$_{0.003}$ & 0.845$_{0.002}$ & 0.868$_{0.005}$ & 0.881$_{0.002}$ & 0.881$_{0.002}$ & 0.884$_{0.004}$ &0.883$_{0.002}$ & 0.905$_{0.003}$ & 0.896$_{0.004}$ & \textbf{0.908$_{0.002}$} \\
& 1-OE & 0.327$_{0.010}$ & 0.410$_{0.007}$ & 0.295$_{0.014}$ & 0.391$_{0.009}$ & 0.311$_{0.015}$ & 0.350$_{0.009}$ & 0.468$_{0.007}$ & 0.453$_{0.006}$ & 0.475$_{0.018}$ & 0.491$_{0.012}$ & \textbf{0.523$_{0.004}$} \\
& 1-Cov & 0.648$_{0.006}$ & 0.723$_{0.007}$ & 0.647$_{0.005}$ & 0.718$_{0.006}$ & 0.702$_{0.008}$ & 0.725$_{0.005}$ & 0.727$_{0.017}$ & 0.727$_{0.006}$ & 0.771$_{0.005}$ & 0.749$_{0.009}$ & \textbf{0.770$_{0.004}$} \\
\midrule
\multirow{6}{*}{Pascal07} 
& AP & 0.496$_{0.004}$ & 0.508$_{0.005}$ & 0.471$_{0.008}$ & 0.504$_{0.005}$ & 0.437$_{0.005}$ & 0.488$_{0.003}$ & 0.505$_{0.012}$ & 0.519$_{0.006}$ & 0.548$_{0.008}$ & 0.551$_{0.004}$ & \textbf{0.564$_{0.005}$} \\
& 1-HL & 0.927$_{0.000}$ & 0.931$_{0.001}$ & 0.928$_{0.001}$ & 0.930$_{0.000}$ & 0.882$_{0.004}$ & 0.928$_{0.001}$ & 0.929$_{0.001}$ & 0.924$_{0.003}$ & 0.931$_{0.001}$ & 0.934$_{0.001}$ & \textbf{0.935$_{0.002}$} \\
& 1-RL & 0.761$_{0.004}$ & 0.812$_{0.004}$ & 0.761$_{0.005}$ & 0.806$_{0.003}$ & 0.736$_{0.015}$ & 0.783$_{0.001}$ & 0.783$_{0.008}$ & 0.811$_{0.004}$ & 0.831$_{0.004}$ & 0.832$_{0.004}$ & \textbf{0.837$_{0.004}$} \\
& AUC & 0.786$_{0.003}$ & 0.838$_{0.003}$ & 0.779$_{0.004}$ & 0.832$_{0.002}$ & 0.767$_{0.015}$ & 0.811$_{0.001}$ & 0.809$_{0.006}$ & 0.834$_{0.004}$ & 0.851$_{0.004}$ & 0.851$_{0.003}$ & \textbf{0.860$_{0.005}$} \\
& 1-OE & 0.434$_{0.005}$ & 0.419$_{0.008}$ & 0.410$_{0.008}$ & 0.419$_{0.006}$ & 0.362$_{0.023}$ & 0.421$_{0.006}$ & 0.427$_{0.015}$ & 0.421$_{0.006}$ & 0.461$_{0.013}$ & 0.459$_{0.007}$ & \textbf{0.474$_{0.020}$} \\
& 1-Cov & 0.703$_{0.004}$ & 0.759$_{0.003}$ & 0.692$_{0.004}$ & 0.751$_{0.003}$ & 0.677$_{0.015}$ & 0.727$_{0.002}$ & 0.731$_{0.006}$ & 0.763$_{0.005}$ & 0.783$_{0.004}$ & 0.784$_{0.003}$ & \textbf{0.790$_{0.005}$} \\
\midrule
\multirow{6}{*}{ESPGame} 
& AP & 0.221$_{0.002}$ & 0.289$_{0.003}$ & 0.212$_{0.002}$ & 0.285$_{0.003}$ & 0.244$_{0.005}$ & 0.246$_{0.002}$ & 0.297$_{0.002}$ & 0.294$_{0.004}$ & 0.305$_{0.004}$ & 0.306$_{0.003}$ & \textbf{0.326$_{0.003}$} \\
& 1-HL & 0.982$_{0.000}$ & 0.983$_{0.000}$ & 0.982$_{0.000}$ & 0.983$_{0.000}$ & 0.972$_{0.002}$ & 0.983$_{0.000}$ & 0.983$_{0.000}$ & 0.982$_{0.000}$ & 0.983$_{0.001}$ & 0.983$_{0.000}$ & \textbf{0.983$_{0.000}$} \\
& 1-RL & 0.780$_{0.004}$ & 0.832$_{0.001}$ & 0.781$_{0.001}$ & 0.822$_{0.008}$ & 0.808$_{0.002}$ & 0.818$_{0.002}$ & 0.838$_{0.001}$ & 0.828$_{0.002}$ & 0.846$_{0.002}$ & 0.831$_{0.003}$ & \textbf{0.848$_{0.001}$} \\
& AUC & 0.784$_{0.004}$ & 0.836$_{0.001}$ & 0.785$_{0.001}$ & 0.833$_{0.002}$ & 0.813$_{0.002}$ & 0.824$_{0.002}$ & 0.836$_{0.001}$ & 0.833$_{0.002}$ & 0.850$_{0.004}$ & 0.842$_{0.002}$ & \textbf{0.859$_{0.002}$} \\
& 1-OE & 0.317$_{0.010}$ & 0.389$_{0.004}$ & 0.313$_{0.009}$ & 0.389$_{0.004}$ & 0.343$_{0.013}$ & 0.339$_{0.003}$ & 0.439$_{0.005}$ & 0.434$_{0.009}$ & 0.442$_{0.011}$ & 0.447$_{0.009}$ & \textbf{0.480$_{0.005}$} \\
& 1-Cov & 0.496$_{0.006}$ & 0.574$_{0.004}$ & 0.488$_{0.005}$ & 0.567$_{0.005}$ & 0.548$_{0.004}$ & 0.571$_{0.003}$ & 0.593$_{0.003}$ & 0.590$_{0.004}$ & 0.624$_{0.005}$ & 0.602$_{0.004}$ & \textbf{0.633$_{0.005}$} \\
\midrule
\multirow{6}{*}{IAPRTC12} 
& AP & 0.256$_{0.002}$ & 0.305$_{0.004}$ & 0.234$_{0.003}$ & 0.304$_{0.004}$ & 0.237$_{0.003}$ & 0.261$_{0.003}$ & 0.323$_{0.003}$ &0.317$_{0.003}$ & 0.329$_{0.005}$ & 0.332$_{0.003}$ & \textbf{0.360$_{0.003}$} \\
& 1-HL & 0.980$_{0.000}$ & 0.981$_{0.000}$ & 0.980$_{0.000}$ & 0.981$_{0.000}$ & 0.969$_{0.002}$ & 0.980$_{0.000}$ & 0.981$_{0.000}$ & 0.980$_{0.000}$ & 0.981$_{0.001}$ & 0.981$_{0.002}$ & \textbf{0.981$_{0.000}$} \\
& 1-RL & 0.832$_{0.004}$ & 0.862$_{0.002}$ & 0.823$_{0.003}$ & 0.861$_{0.002}$ & 0.833$_{0.002}$ & 0.844$_{0.003}$ & 0.873$_{0.001}$ & 0.870$_{0.001}$ & 0.883$_{0.003}$ & 0.875$_{0.002}$ & \textbf{0.888$_{0.005}$} \\
& AUC & 0.830$_{0.001}$ & 0.864$_{0.002}$ & 0.825$_{0.003}$ & 0.863$_{0.003}$ & 0.835$_{0.001}$ & 0.850$_{0.001}$ & 0.874$_{0.002}$ & 0.872$_{0.001}$ & 0.885$_{0.003}$ & 0.876$_{0.001}$ & \textbf{0.896$_{0.001}$} \\
& 1-OE & 0.430$_{0.008}$ & 0.429$_{0.005}$ & 0.352$_{0.008}$ & 0.429$_{0.005}$ & 0.314$_{0.009}$ & 0.386$_{0.003}$ & 0.468$_{0.007}$ & 0.443$_{0.005}$ & 0.459$_{0.080}$ & 0.467$_{0.005}$ & \textbf{0.507$_{0.007}$} \\
& 1-Cov & 0.534$_{0.003}$ & 0.597$_{0.004}$ & 0.529$_{0.004}$ & 0.597$_{0.004}$ & 0.564$_{0.005}$ & 0.592$_{0.004}$ & 0.649$_{0.004}$ & 0.648$_{0.003}$ & 0.673$_{0.001}$ & 0.649$_{0.004}$ & \textbf{0.688$_{0.007}$} \\
\midrule
\multirow{6}{*}{MIRFlickr} 
& AP & 0.537$_{0.005}$ & 0.570$_{0.006}$ & 0.514$_{0.004}$ & 0.553$_{0.004}$ & 0.490$_{0.010}$ & 0.510$_{0.005}$ & 0.589$_{0.005}$ & 0.594$_{0.005}$ & 0.602$_{0.009}$ & 0.607$_{0.004}$ & \textbf{0.615$_{0.001}$} \\
& 1-HL & 0.874$_{0.001}$ & 0.888$_{0.001}$ & 0.882$_{0.001}$ & 0.885$_{0.001}$ & 0.839$_{0.002}$ & 0.882$_{0.001}$ & 0.888$_{0.001}$ & 0.882$_{0.002}$ & 0.890$_{0.001}$ & 0.891$_{0.001}$ & \textbf{0.894$_{0.001}$} \\
& 1-RL & 0.832$_{0.001}$ & 0.844$_{0.002}$ & 0.844$_{0.002}$ & 0.856$_{0.004}$ & 0.803$_{0.004}$ & 0.844$_{0.002}$ & 0.830$_{0.007}$ & 0.863$_{0.004}$ & 0.883$_{0.003}$ & 0.875$_{0.002}$ & \textbf{0.890$_{0.004}$} \\
& AUC & 0.817$_{0.002}$ & 0.844$_{0.001}$ & 0.830$_{0.001}$ & 0.844$_{0.001}$ & 0.787$_{0.003}$ & 0.837$_{0.001}$ & 0.849$_{0.002}$ & 0.853$_{0.003}$ & 0.861$_{0.001}$ & 0.862$_{0.001}$ & \textbf{0.873$_{0.003}$} \\
& 1-OE & 0.511$_{0.012}$ & 0.579$_{0.002}$ & 0.517$_{0.008}$ & 0.631$_{0.002}$ & 0.490$_{0.010}$ & 0.510$_{0.005}$ & 0.652$_{0.007}$ & 0.642$_{0.008}$ & 0.651$_{0.006}$ & 0.655$_{0.004}$ & \textbf{0.672$_{0.005}$} \\
& 1-Cov & 0.605$_{0.004}$ & 0.649$_{0.003}$ & 0.597$_{0.003}$ & 0.636$_{0.004}$ & 0.572$_{0.013}$ & 0.631$_{0.002}$ & 0.652$_{0.007}$ & 0.667$_{0.003}$ & 0.671$_{0.004}$ & 0.676$_{0.004}$ & \textbf{0.690$_{0.002}$} \\
\bottomrule
\end{tabular}
}
\end{table*}

\begin{figure*}[!t]
    \centering
    \subfloat[Corel5k]{%
        \includegraphics[width=0.31\linewidth,height=1.8in]{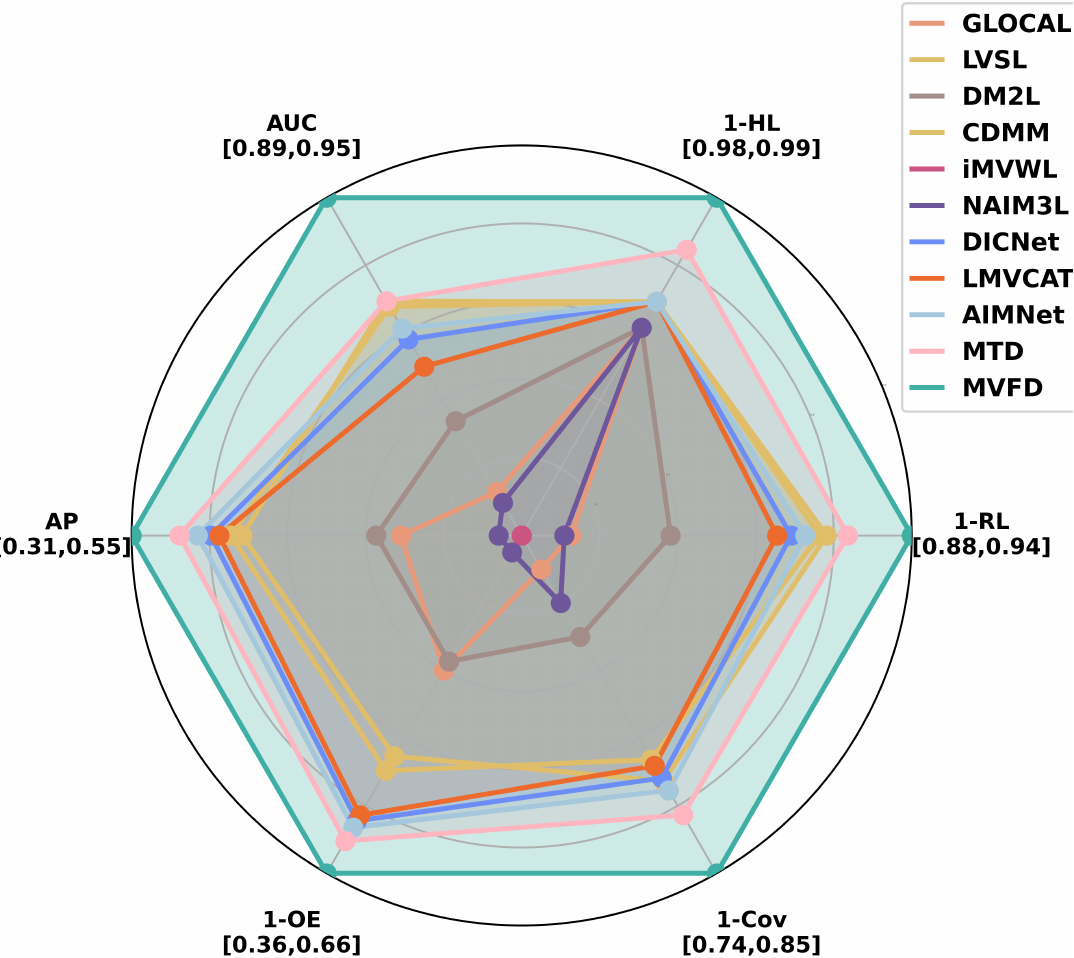}
        }
    \hfill
    \subfloat[Pascal07]{%
        \includegraphics[width=0.31\linewidth,height=1.8in]{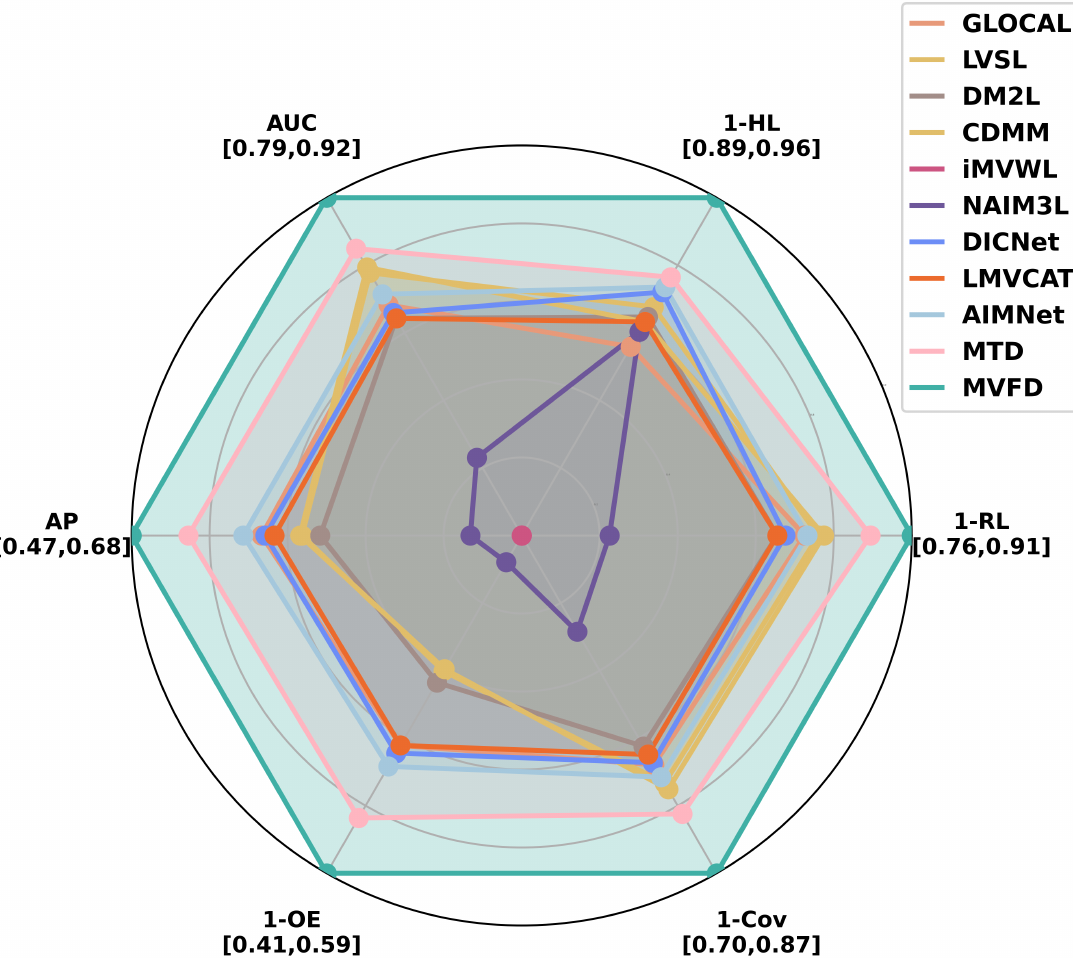}
        }
    \hfill
    \subfloat[ESPGame]{%
        \includegraphics[width=0.31\linewidth,height=1.8in]{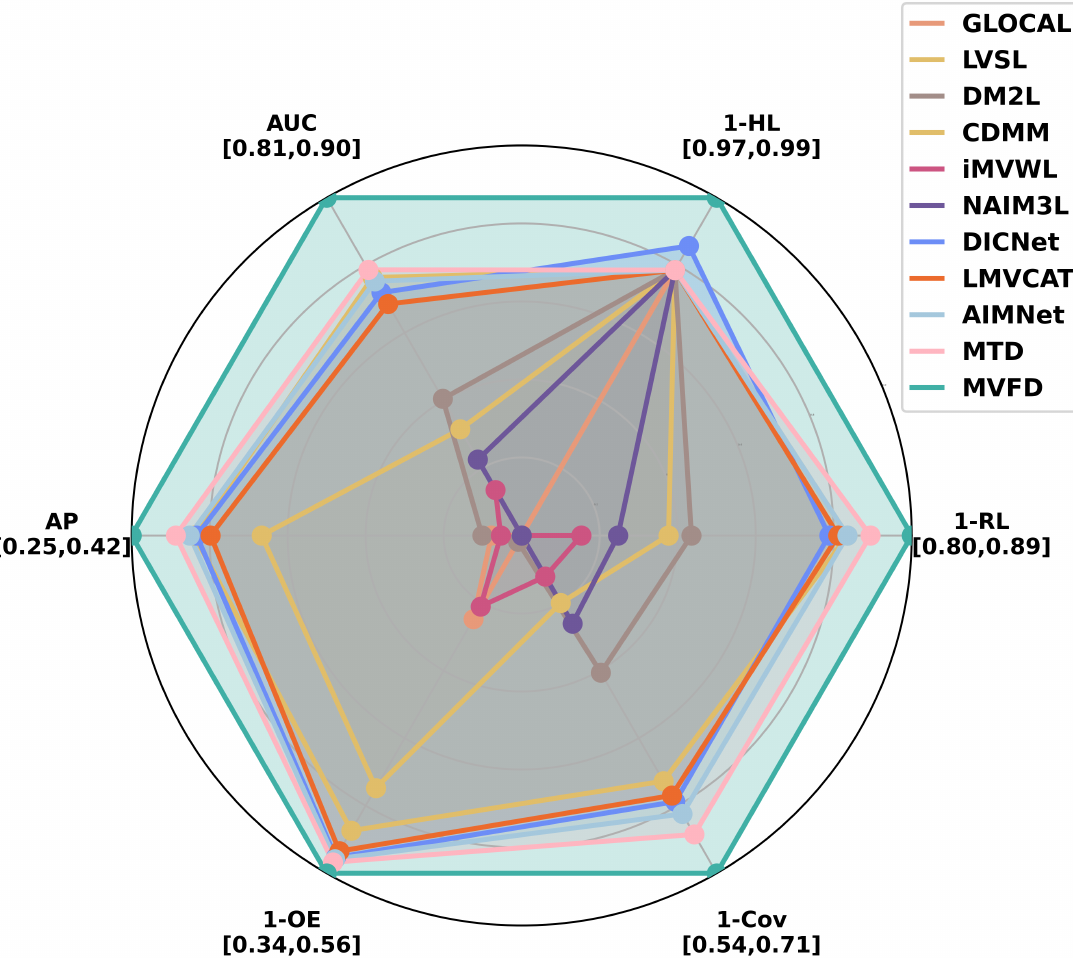}
        }
    \caption{Experimental results of eleven methods on three datasets without any missing views or labels. The worst results are indicated at the center of the radar chart while the best results are represented by the vertexes on the six metrics.}
\end{figure*}

\begin{figure}[!t]
\small
\centering
\subfloat[Missing on View]{\includegraphics[width=1.6in,height=1.2in]{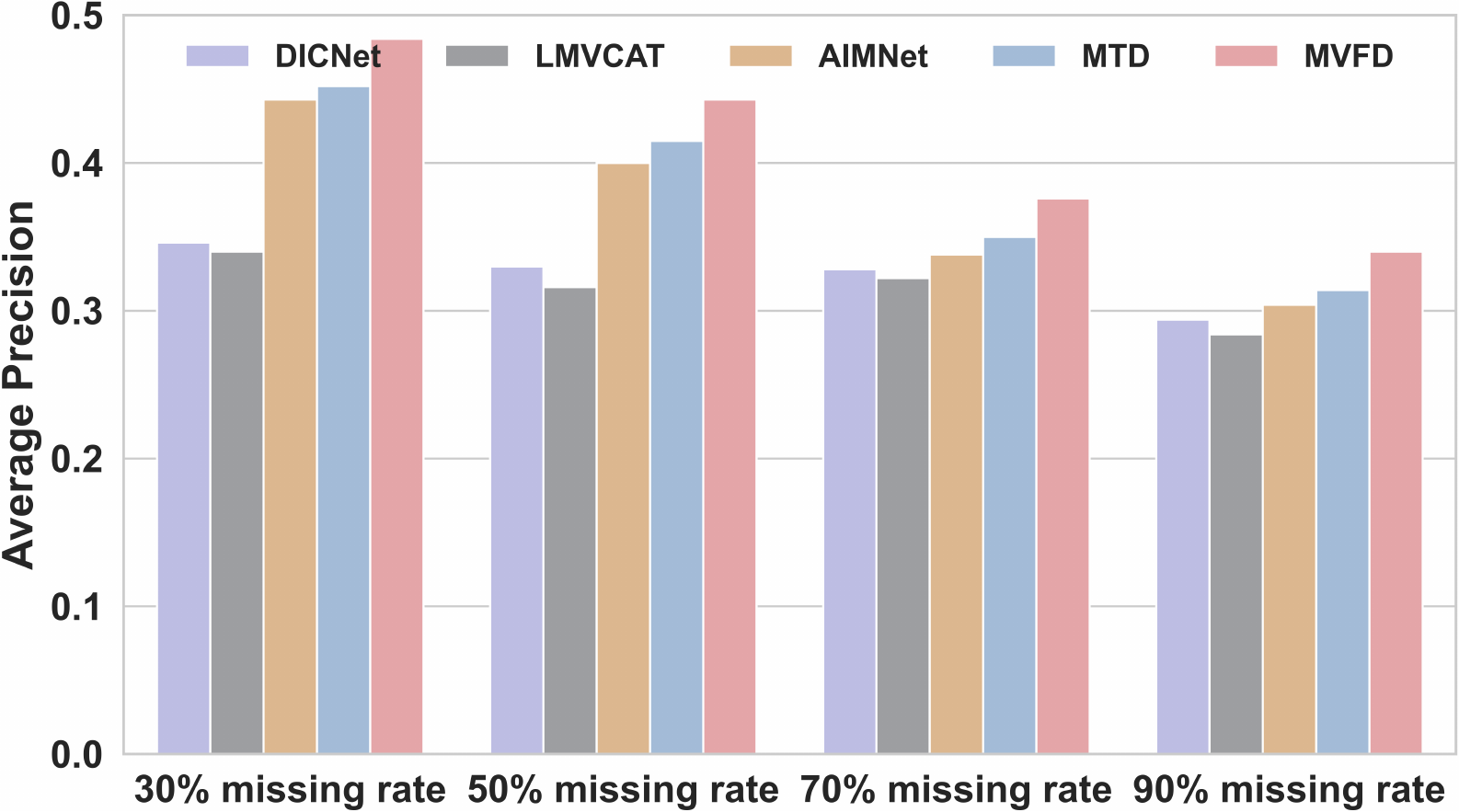}}
\hfil
\subfloat[Missing on Label]{\includegraphics[width=1.6in,height=1.2in]{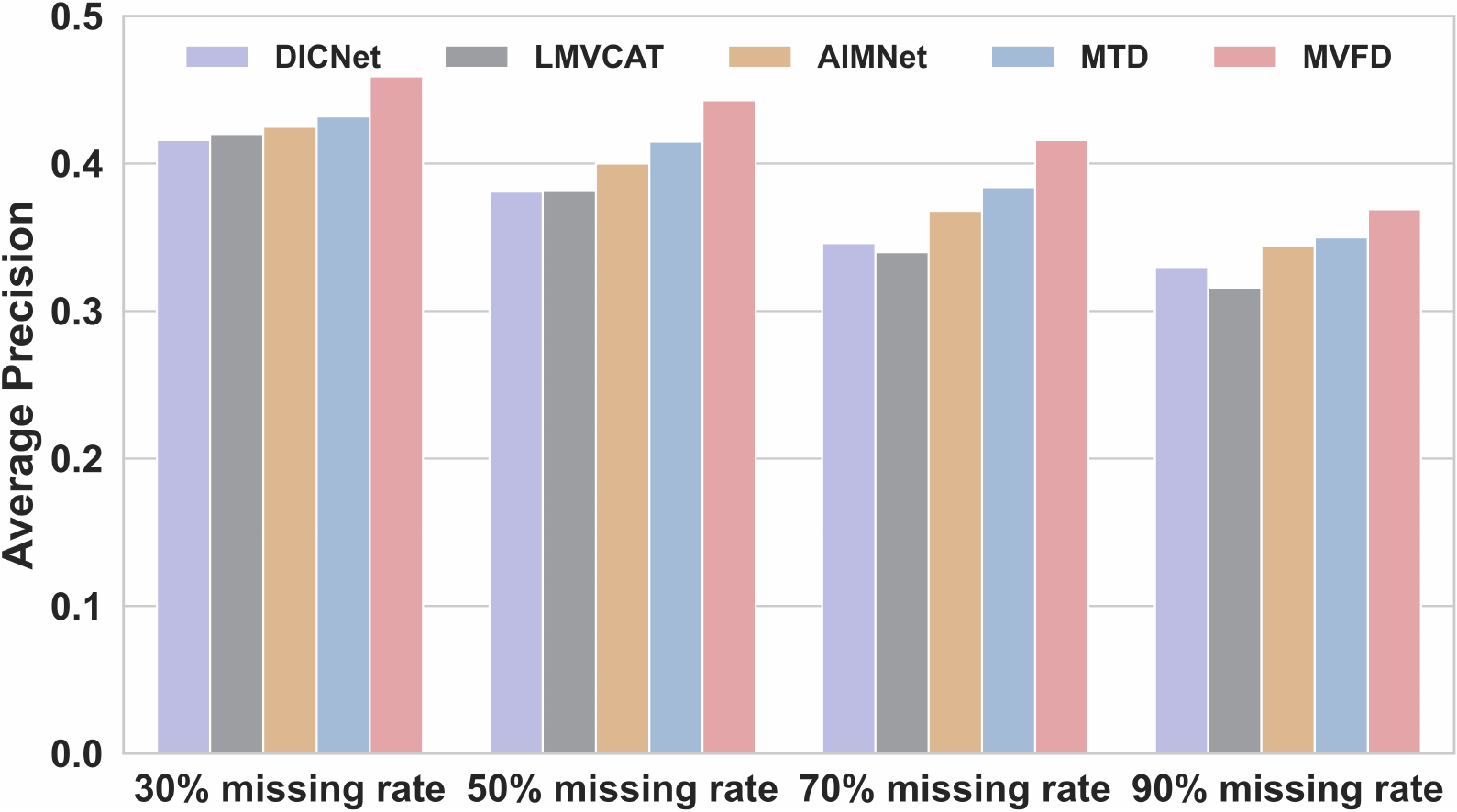}}
\caption{Performance about the comparison with SOTA methods with different missing rates on View or Label}
\vspace{-3mm} 
\end{figure}

\begin{figure}[!t]
\small
\centering
\subfloat[Corel5k]{\includegraphics[width=1.6in,height=1.2in]{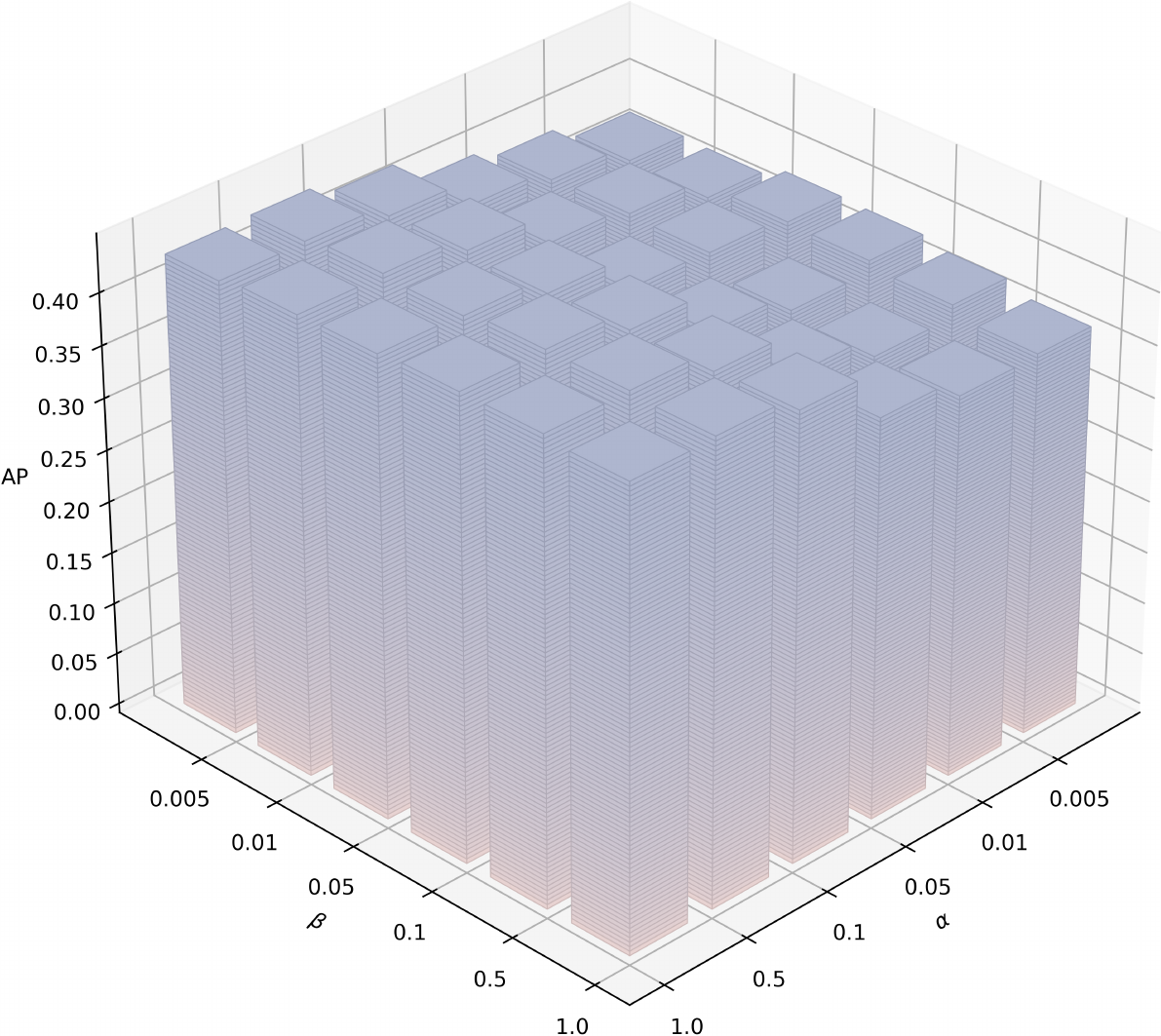}}
\hfil
\subfloat[Pascal07]{\includegraphics[width=1.6in,height=1.2in]{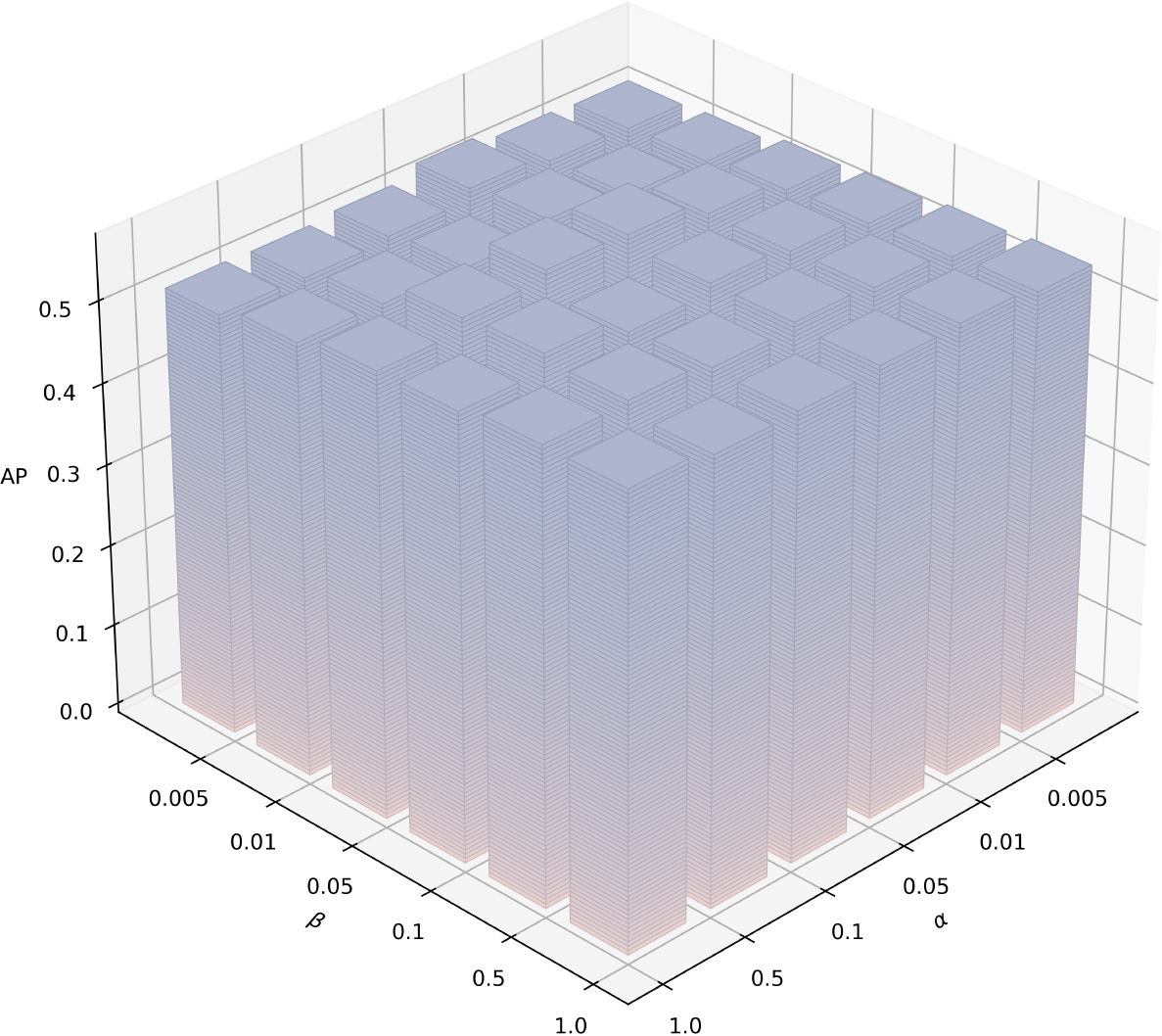}}
\hfil
\subfloat[Corel5k]{\includegraphics[width=1.6in,height=1.2in]{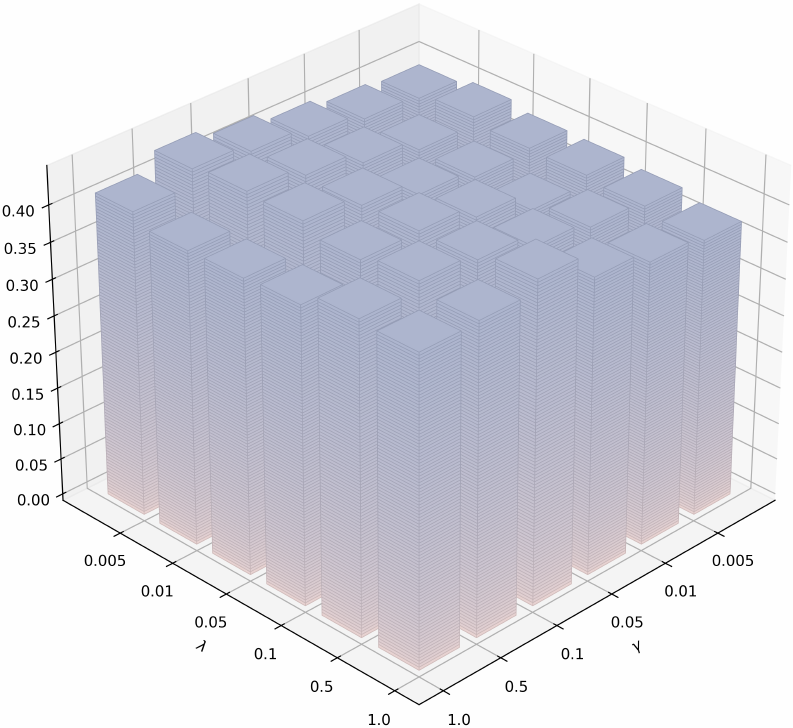}}
\hfil
\subfloat[Pascal07]{\includegraphics[width=1.6in,height=1.2in]{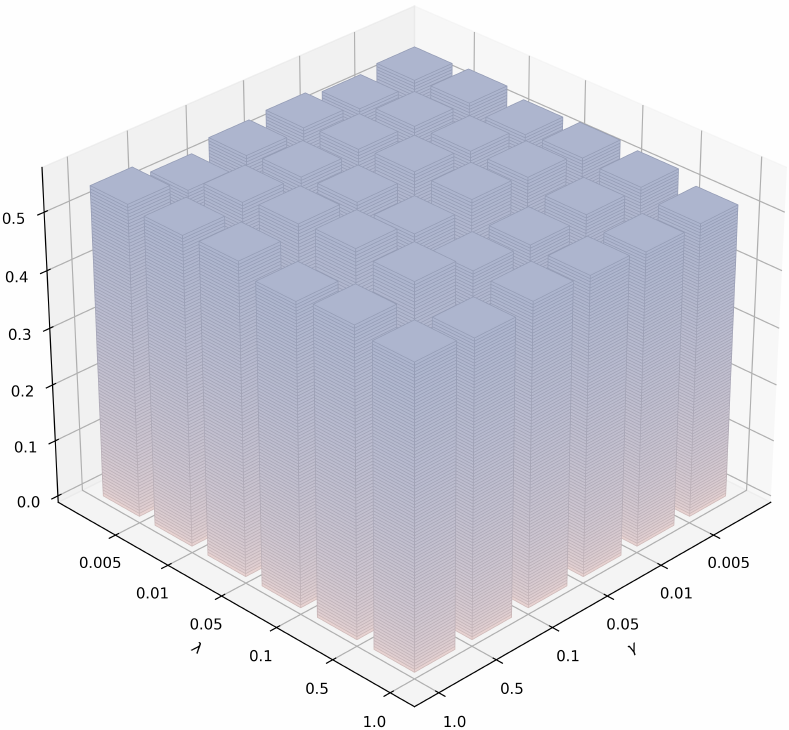}}
\hfil
\caption{The AP values for hyper-parameters $\alpha$ and $\beta$ on the Corel5k (Fig5. a) and Pascal07 (Fig5. b) datasets; AP values for hyper-parameters $\alpha$ and $\beta$ on the Corel5k (Fig5. c) and Pascal07 (Fig5. d) datasets are presented. Both datasets contain 50\% available views and labels, with a 70\% training sample rate.}
\vspace{-4mm} 
\end{figure}

\begin{figure}[!t]
\small
\centering
\subfloat[Epoch 1]{\includegraphics[width=1.2in,height=1.2in]{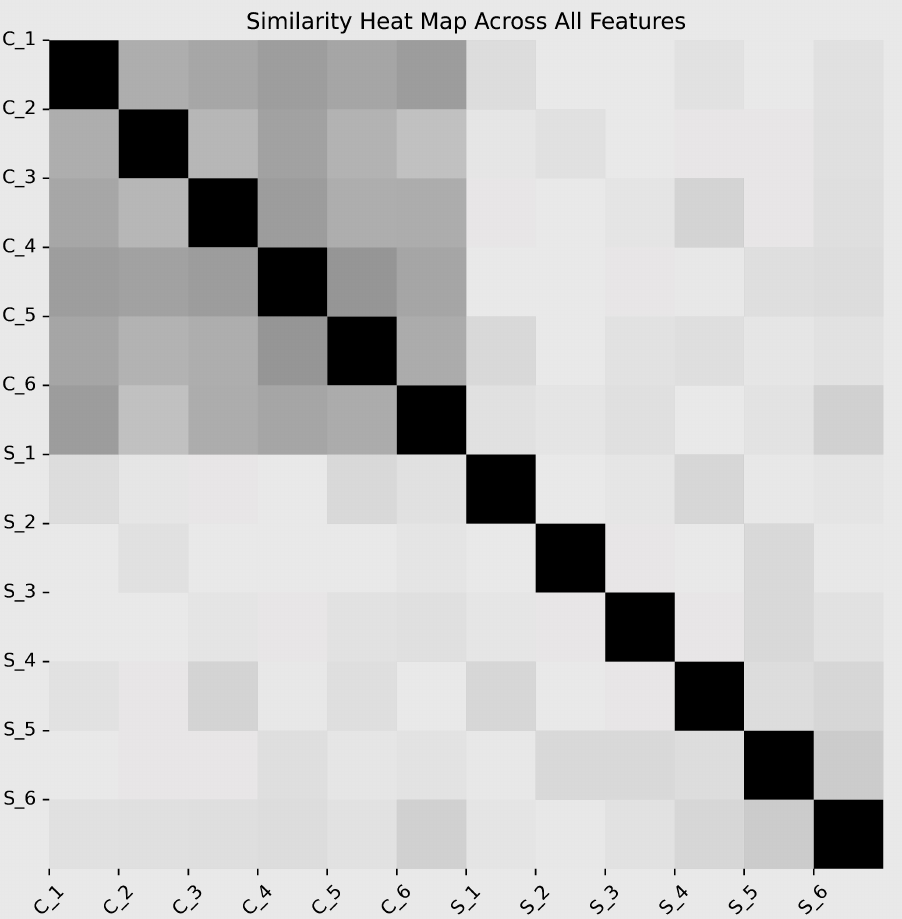}}
\hfil
\subfloat[Epoch 20]{\includegraphics[width=1.2in,height=1.2in]{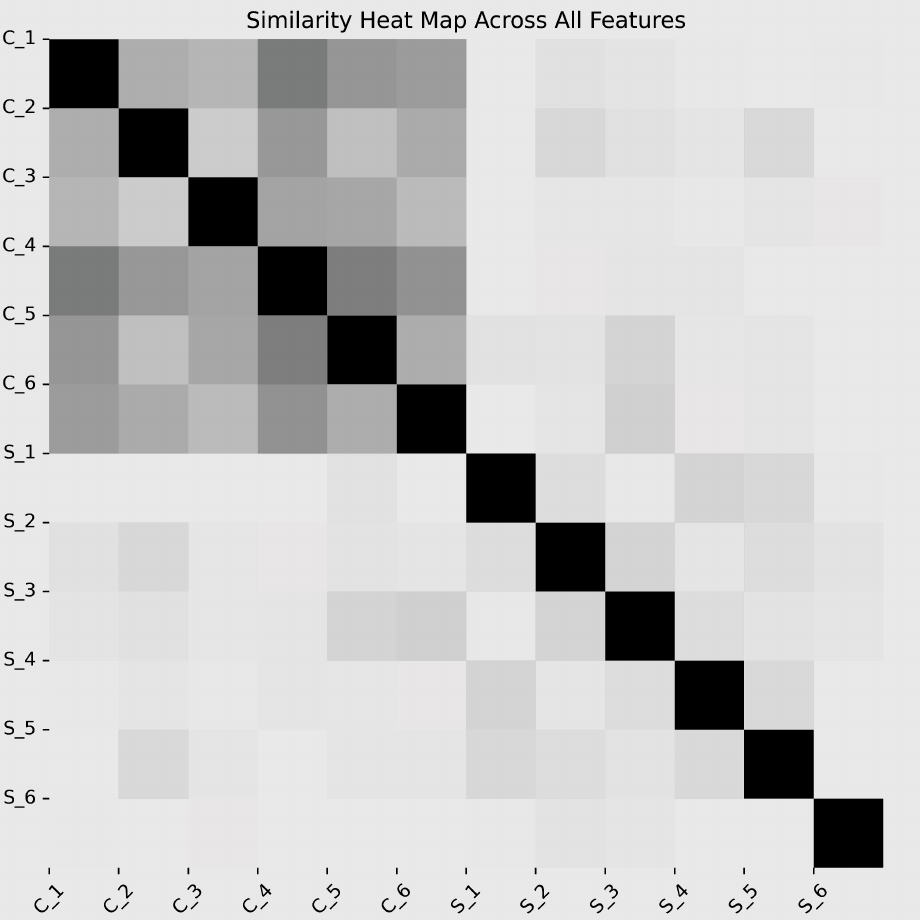}}
\hfil
\subfloat[Epoch 40]{\includegraphics[width=1.2in,height=1.2in]{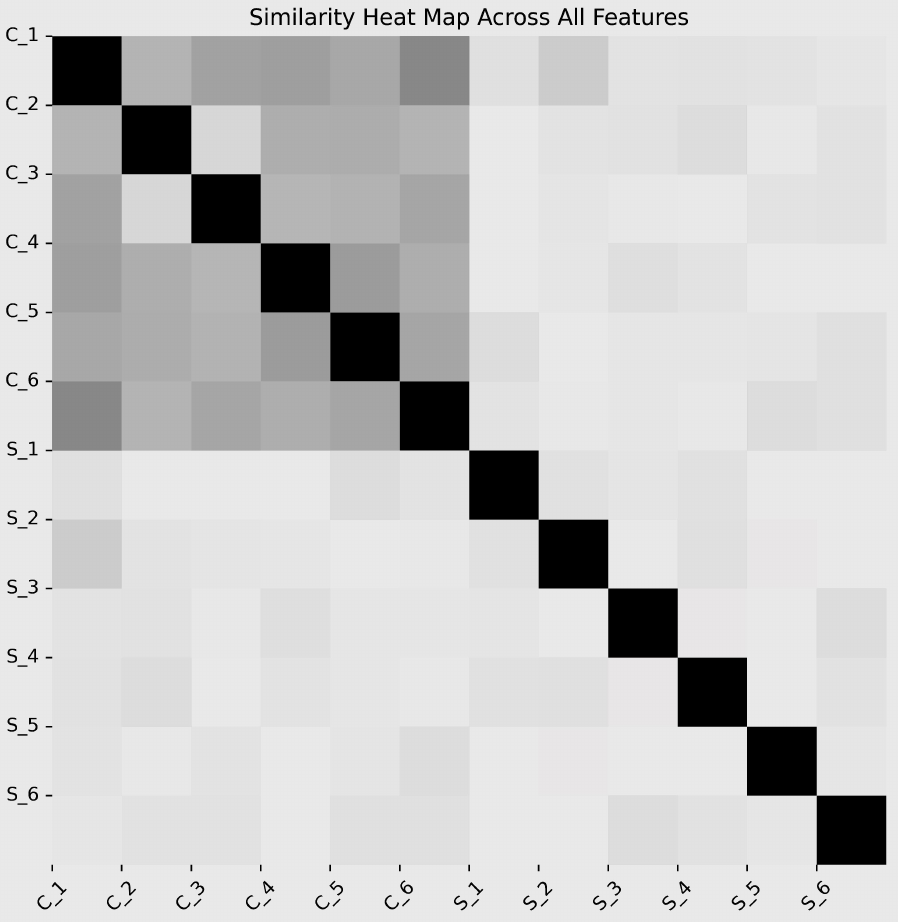}}
\hfil
\subfloat[Epoch 60]{\includegraphics[width=1.2in,height=1.2in]{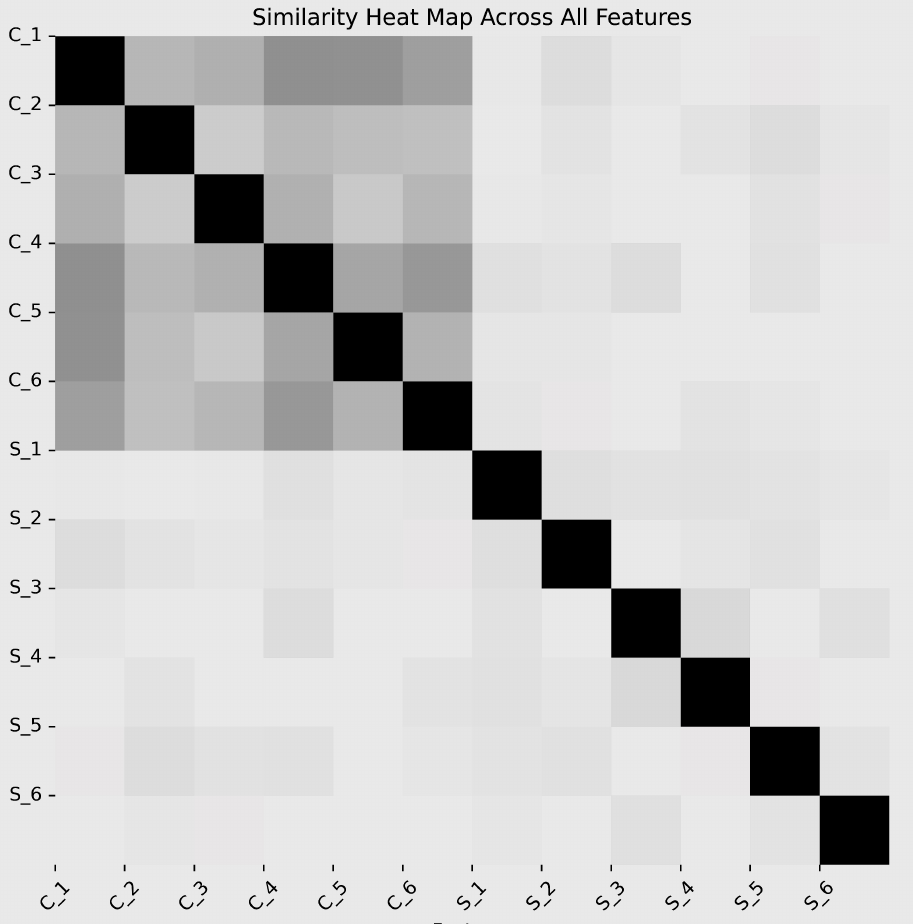}}
\hfil
\caption{A random sample’s feature similarity heat maps across different informations on Corel5k dataset.}
\vspace{-4mm} 
\end{figure}

\subsection{Experimental Results and Analysis}
To prove the effectiveness of our method, we compare our MVFD with ten top methods on five datasets with a 50\% missing rate in both views and labels, where Table 1 shows the experimental results of the six evaluation metrics. Based on the results in Table 1, we can have the following observations:
\begin{itemize}
\item On all five datasets, our MVFD outperforms comparison methods on all metrics, which fully verifies the effectiveness of our method.
\item From the results shown in Table 1, we can observe that DNN-based methods, such as DICNet, LMVCAT, AIMNet, MTD and our MVFD, exhibit better performance than other methods, which shows the potential of deep neural network in iMvMLC task.
\item The methods such as DICNet and our MVFD that consider the incompleteness problem in both views and labels, have better performance compared with other methods that either ignore the incompleteness problem in both view and label or only consider single missing problem
\end{itemize}
Additionally, to further confirm that our model has good adaptability to complete multi-view data, we provide the results of nine methods with full views and labels on Corel5k, Pascal07 and EPSGame dataset in Figure. 3, which can be observed that our MVFD still achieve competitive performance than other methods, including those designed for the ideal complete case, demonstrating the generalization ability of our model. 

Moreover, to provide a more comprehensive demonstration of
our model in handling missing values, we present the experimental results with different missing rates on view and label on Corel5K dataset. As shown in Fig .4a and Fig. 4b, proposed MVFD can achieve best AP score with all missing settings on View or Label.

\subsection{Hyperparameters Study}
The overall loss of our model mainly contains 4 hyperparameters, i.e., $\alpha$, $\beta$, $\gamma$ and $\lambda$. Specially, $\alpha$ and $\beta$ are the weight coefficients of $\ell _{cp} $ and $\ell _{sc} $ in the first stage,  $\gamma$ and $\lambda$ are the weight coefficients of $\ell _{rec} $ and $\ell _{gd} $ in the second stage. To study the impact of different hyperparameters on the performance of our model, we iteratively set different values and report the corresponding AP values on Corel5k and Pascal07 dataset, with 50\% missing view and labels, with 70\% training samples. Fig. 5a and Fig. 5b show the AP value versus hyperparameters $\alpha$ and $\beta$. Fig .5c and Fig .5d show the AP values versus hyperparameters $\gamma$ and $\lambda$. As shown in Fig 5a and Fig 5b, the optimal ranges for $\alpha$ and $\beta$ on Corel5k are [0.5,1] and [0.05,0.5], and on Pascal07, they are both [0.05,0.1]. From Fig .5c and Fig. 5d, the the optimal ranges for $\gamma$ and $\lambda$ on Corel5k are [0.1,0.5] and [0.005,0.01], and on Pascal07, they are [0.05,1] and [0.05,0.5], respectively. 

\subsection{Analysis of Disentangment}
To demostrate the effectiveness of our graph disentangling loss, inspired by \cite{liu2024masked}, we plot a random sample’s channel similarity heat maps across view-shared feature and view-specific feature in different training epochs in stage 2 on the Corel5k dataset with half of missing views and labels. For any sample $i$, we obtain $\mathbf{O}_{i}$ by concatating its view-shared feature $\left\{\mathbf{C}_{i}^{(v)}\right\}_{v=1}^{m}$ with its view-specific feature $\left\{\mathbf{S}_{i}^{(v)}\right\}_{v=1}^{m}$, then we can calculate the $i$-th sample's heat map $\mathbf{H}_{i} = \mathbf{O}_{i}\mathbf{O}_{i}^{T}$ where $\mathbf{H}_{i}$ shows the cosine similarity of sample $i$'s different features.

From Fig. 6(a), we can observe that in the initial state of stage 2, the similarities for the view-shared features are high, which demonstrate the effectiveness of our factorized consistent representation learning strategy in stage 1. From Fig. 6(b)(c)(d), it is evident that as training progresses, The similarity between view-consistent features and view-specific features, as well as between different view specific features, is gradually decrease, which indicates that our graph disentangling loss is functioning as expected.

\begin{table}[H]
\footnotesize
\centering
\caption{The ablation experiment result on Corel5k dataset and Pascal07 dataset.}
\label{my-label}
\resizebox{8.5cm}{!}{
\begin{tabular}{@{}ccccc|cc|cc@{}}
\toprule
Backbone & $L_{cp}$ & $L_{sc}$ & $L_{rec}$ & $L_{gd}$ & \multicolumn{2}{c}{Corel5k} & \multicolumn{2}{c}{Pascal07} \\
 &  &  &  &  & AP & AUC & AP & AUC \\
\midrule
$\checkmark$ &  &  &  &  & 0.398 & 0.890 & 0.530 & 0.820 \\
$\checkmark$ & $\checkmark$ &  &  &  & 0.416 & 0.899 & 0.545 & 0.845 \\
$\checkmark$ &  & $\checkmark$ &  &  & 0.415 & 0.897 & 0.543 & 0.844 \\
$\checkmark$ &  &  & $\checkmark$ &  & 0.403 & 0.892 & 0.534 & 0.833 \\
$\checkmark$ &  &  &  & $\checkmark$ & 0.406 & 0.898 & 0.538 & 0.840 \\
$\checkmark$ & $\checkmark$ & $\checkmark$ &  &  & 0.430 & 0.904 & 0.565 & 0.851 \\
$\checkmark$ &  &  & $\checkmark$ & $\checkmark$  & 0.429 & 0.905 & 0.563 & 0.854 \\
$\checkmark$ & $\checkmark$ & $\checkmark$ & $\checkmark$ & $\checkmark$ & 0.441 & 0.908 & 0.564 & 0.860 \\
\midrule
\multicolumn{5}{c}{one stage} & 0.426 & 0.900 & 0.547 & 0.845 \\
\bottomrule
\end{tabular}
}
\end{table}

\subsection{Ablation Study}
To validate each module's effectiveness, we conduct ablation experiments on Corel5k and Pascal07 datasets with a 50\% missing rate of views and labels. Loss functions $\ell _{cp} $, $\ell _{sc} $, $\ell _{rec} $, and $\ell _{gd} $ are sequentially removed, while the backbone retains classification losses $\ell _{ce1} $ and $\ell _{ce2} $ in two stages. To compare the two-stage learning paradigm with an end-to-end framework, we consolidate the two-stage model into an end-to-end one by moving first-stage losses to the second stage and training all modules simultaneously. Results in Table 2 show: (i): All designs introduced into our method are crucial and result in an enhancement in performance metrics. (ii): When we transfer the two-stage model to the end-to-end model, there is a performance degradation, which confirms that the two-stage learning paradigm may have an advantage in extracting consistent and view-specific information from multi-view data compared with the end-to-end framework.

\section{Conclusion}
In this paper, we propose a novel framework, namely MVFD for incomplete multi-view multi-label classification (iMvMLC) tasks. Specifically, we novelty factorize consistency learning into three sub-objects and design corresponding collaborative consistency learning modules based on masked cross-view prediction (MCP) strategy and information theory. Furthermore, we decompose multi-view representation into view-consist representation and view-specific representation, and proposed a graph disentangling loss to fully eliminate redundant information between them. Finally, we conduct extensive comparison and ablation experiments to demonstrate the effectiveness of our method.


{\small
\bibliographystyle{ieee_fullname}
\bibliography{PaperForReview}
}

\end{document}